\documentclass{article}
\usepackage[preprint]{neurips_2024}
\usepackage[utf8]{inputenc}
\usepackage[T1]{fontenc}
\usepackage{hyperref}
\usepackage{url}
\usepackage{booktabs}
\usepackage{amsfonts}
\usepackage{amsmath}
\usepackage{nicefrac}
\usepackage{microtype}
\usepackage{graphicx}
\usepackage{xcolor}
\usepackage{enumitem}

\title{PlanarBench: Evaluating LLM Spatial Reasoning\\ via Planar Graph Drawing}

\author{%
  Oleksandr Nikitin \\
  \texttt{oleksandr@tvori.info}
  \And
  Anna Kravchenko\\
    \texttt{anna.kravchenko@ru.nl}
}

\begin{document}

\maketitle

\begin{abstract}
Existing LLM graph benchmarks typically ask models to answer graph-theoretic questions or compute symbolic solutions rather than construct spatial layouts. Within-task difficulty is also primarily stratified by vertex count. However, existing research also suggests that task difficulty is more closely related to the number of constraints imposed by the edges than to the number of vertices being arranged. We introduce PlanarBench, a benchmark that asks models to produce crossing-free ASCII drawings of planar graphs given only an edge list. Across 91 model configurations and 199 non-isomorphic connected planar graphs with 2–7 vertices, edge count is more strongly associated with mean task score than vertex count (\(r=-0.85\) versus \(r=-0.47\)) and remains strongly associated after controlling for vertex count (\(r_{\mathrm{partial}}=-0.80\)). PlanarBench provides a controlled setting for separating these two difficulty axes. In addition, neither drawing area nor total response length demonstrated a meaningful correlation with score, which is evidence against a simple output-size explanation. Performance varies widely: the best model scores 159.5 out of 199, most models below 30B parameters score under 25, and substantial failures remain among frontier systems. 
\end{abstract}

\section{Introduction}

Spatial reasoning---including the ability to mentally manipulate objects in two- or three-dimensional space---is a core component of human intelligence that remains poorly tested in LLM benchmarks. Although existing benchmarks do evaluate spatial relations and navigation in LLMs ~\citep{li2024stepgame,wu2024mindseye,premsri2025forest}, construction of verifiable spatial layouts from symbolic graph descriptions remains comparatively underexplored. The major ones instead focus on graph-theoretic queries and algorithmic outputs, such as connectivity, cycle detection, shortest paths, traversal, matching, and optimization. These tasks reduce to symbolic computation and can often be solved by memorizing algorithmic templates.

We propose a fundamentally different evaluation: given a planar graph as an edge list, produce a valid planar straight-line drawing as ASCII art. This task has several desirable properties:

\paragraph{It requires spatial reasoning.}
The model must jointly decide on node positions and verify that all edges can be drawn without crossings. Algorithmic solutions exist---F\'ary's theorem~\citep{fary1948} guarantees every planar graph admits a straight-line embedding, Tutte~\citep{tutte1963} gives a constructive method for 3-connected cases, and de~Fraysseix et al.~\citep{defraysseix1990} provide a general $O(n \log n)$ algorithm producing drawings on an $O(n) \times O(n)$ grid---but the model must implicitly navigate this space without executing a known algorithm.

\paragraph{It is verifiable.}
Unlike open-ended generation tasks, correctness can be checked mechanically by parsing the ASCII art, extracting node coordinates and edge paths, and verifying they match the input graph with no missing or extra edges, and no crossings.

\paragraph{It scales naturally.}
Graph drawing difficulty increases with both vertex count and edge density, providing a built-in difficulty gradient without requiring manual task design.

\paragraph{It probes multi-constraint reasoning.}
Each edge is a constraint that interacts with every other---satisfying one adjacency may block a crossing-free route for another. The model must find a globally consistent assignment, mirroring tasks like checking a specification for contradictory requirements. Our results confirm this: difficulty scales with the number of pairwise interactions (edges, $r = -0.85$), not the number of entities (vertices, $r = -0.47$). Graph drawing may thus serve as a domain-neutral probe for constraint-satisfaction capacity---one that sidesteps training-data contamination because the constraints are geometric and vertex labels are permutable.

\paragraph{Memorization is an unlikely strategy.}

The primary defense is corpus scarcity: while ASCII art is common in training
data, systematic collections of crossing-free ASCII drawings of arbitrary
planar graphs, indexed by edge list, are not a genre that occurs at any
meaningful scale in web text. The secondary barrier is relabeling: vertex labels
can be freely permuted (\texttt{A-B, B-C, C-A} is isomorphic to
\texttt{X-Y, Y-Z, Z-X}), so reusing a memorized drawing requires recognizing
the isomorphism and inverting the permutation. We deliberately do not lean on
the computational hardness of graph isomorphism alone: although isomorphism
testing is quasipolynomial in general~\cite{babai2016} and linear for planar
graphs~\cite{hopcroft1974}, graphs on 2--7 vertices are small enough that
isomorphism recognition is trivial in practice.

\section{Related Work}

As we mentioned in the introduction, work evaluating LLMs on graph tasks has focused on graph computation
rather than graph construction. Wang et al.~\cite{wang2023} introduce NLGraph,
testing LLMs on graph problems (connectivity, shortest path, bipartiteness)
stated in natural language. GraphArena~\cite{tang2025} tests algorithmic tasks
like shortest paths, maximum flow, and NP-complete graph problems.
GraphOmni~\cite{xu2026} evaluates graph-theoretic reasoning across graph types,
serialization formats, and prompt schemes. All of these test
symbolic/algorithmic reasoning over graphs, none of them requires the model to produce
a spatial layout.
 
In graph visualization, Fan et al.~\cite{fan2025} evaluate LLM-generated
layouts for graph understanding, layout generation, and layout evaluation, but
do not test strict planarity. Di Bartolomeo et al.~\cite{dibartolomeo2023} test
whether LLMs can execute steps of graph layout algorithms such as
Sugiyama-style hierarchical layout, but the task is procedural rather than
requiring a holistic layout.
 
Treating edge count as an object of analysis is also uncommon. Difficulty is
typically reported and stratified by vertex count, with edge density varying as
an uncontrolled property of the sampled instances.

Prior work nevertheless indicates that vertex count alone is an incomplete measure of graph-task difficulty. The Scaled Graph Reasoning benchmark varies graph scale through edge count at fixed density and finds that model performance does not change monotonically with graph size \citep{ge2024}. In graph coloring, every edge imposes a pairwise inequality constraint, and Heyman and Zylberberg find that accuracy generally declines as edge count increases within fixed vertex- and color-count conditions, although this relationship can reverse in dense regimes where instances become uniformly uncolorable \citep{heyman2025}. These studies identify edge structure as a relevant difficulty factor, but they do not directly compare edge count and vertex count as competing predictors. PlanarBench makes that comparison explicitly.

\section{Benchmark Design}

\subsection{Task Formulation}

Each benchmark instance consists of:
\begin{itemize}[nosep]
  \item \textbf{Input:} A planar graph specified as an edge list (e.g., \texttt{A - B, A - C, B - C, C - D}).
  \item \textbf{Prompt:} \texttt{"this is a graph: [edge list]. draw an ascii art representation of it, enclosed in a code block. avoid intersections, this is a planar graph."}
  \item \textbf{Expected output:} A code block containing ASCII art where each node appears exactly once as a single letter at a distinct position; edges are drawn using \texttt{-} (horizontal), \texttt{|} (vertical), \texttt{/} (diagonal up), or \texttt{\textbackslash} (diagonal down); all edges from the input graph are present; and no spurious edges are detected.
\end{itemize}

\subsection{Graph Corpus}

Our graphs are a prefix of the complete catalog of all non-isomorphic connected planar graphs, generated by \texttt{plantri}~\citep{brinkmann2007,mckay2024} and stored in graph6 format~\citep{mckay2014}. The evaluated set includes all such graphs on 2--6 vertices and the first 71 of 646 on 7 vertices, for 199 graphs total (Table~\ref{tab:vertex-dist}).

\begin{table}[h]
\centering
\caption{Distribution of benchmark graphs by vertex count.}
\label{tab:vertex-dist}
\small
\begin{tabular}{@{}lrr@{}}
\toprule
Vertices & Count & Cumulative \\
\midrule
2 & 1 & 1 \\
3 & 2 & 3 \\
4 & 6 & 9 \\
5 & 20 & 29 \\
6 & 99 & 128 \\
7 & 71 & 199 \\
\bottomrule
\end{tabular}
\end{table}

We did not extend beyond 7 vertices because the current set already saturates model capability: even with relaxed scoring, solve rates for 11- and 12-edge tasks remain below 2\%, and 8-vertex graphs have up to 18 edges---well past this effective ceiling. Adding larger graphs would increase the corpus without adding discriminative power at current capability levels.

\subsection{Validation Pipeline}

The validator operates in three stages:

\begin{enumerate}[nosep]
\item \textbf{Node extraction.} All \verb|\b\w+\b| tokens in the code block are extracted as node names. The validator checks for missing nodes, extra nodes, and duplicates.

\item \textbf{Edge detection.} The code block is parsed into a character grid. Node positions are identified, then edges are detected in priority order: \emph{horizontal} (same-row nodes connected by contiguous dash characters), \emph{vertical} (same-column nodes connected by pipe characters), and \emph{diagonal} (nodes in a \texttt{\textbackslash} or \texttt{/} relationship). Detected edge characters are ``consumed'' (replaced with a numeric marker) to prevent double-counting.

\item \textbf{Edge matching.} Detected edges are compared against the expected edge set. A drawing passes \emph{strict} validation only if both sets match exactly---no missing edges, no extra edges, and no unconsumed edge characters remain.
\end{enumerate}

Because strict straight-line validation penalizes drawings that are topologically correct but use L- or C-shaped edge paths (common when ASCII grid constraints make straight lines impossible), we additionally run two relaxed checks:
\begin{itemize}[nosep]
\item \textbf{Coordinate validation} checks that all desired edges can be drawn as straight lines without intersections given the node positions, and rejects extra nodes and detectable spurious edges.
\item \textbf{BFS validation} checks edge connectivity by running breadth-first search between adjacent nodes, accepting any contiguous path of edge characters (not just straight lines).
\end{itemize}
The per-task score is $\min\!\bigl(1,\;\max(\text{strict},\;\tfrac{1}{2}\text{coords} + \tfrac{1}{2}\text{bfs})\bigr)$: full credit for a correct straight-line drawing, half credit each for correct node placement and correct (possibly non-straight) edge connectivity, and zero otherwise.

\section{Experimental Setup}

\subsection{Models}

We evaluate 91 models released from December 2023 to December 2025, spanning major providers and architectures: Anthropic (Claude Opus~3, 4.5; Sonnet~3.5, 3.7, 4.5; Haiku~3.5, 4.5, with and without extended thinking), OpenAI (GPT-5.2~Pro, GPT-5.2, GPT-5, GPT-5~Mini, GPT-5~Nano, GPT-5~Codex, GPT-4.5~Preview, GPT-4.1/Mini/Nano, GPT-4o/Mini, o1/Mini, o3/Mini, o4~Mini), Google (Gemini~3~Pro/Flash~Preview, Gemini~2.5~Pro, Gemini~2.0~Pro, Gemini~2.0~Flash), xAI (Grok-2, Grok-3/Mini, Grok-4.1~Fast), Moonshot (K2, K2~Thinking), Aion~1.0, Quasar~Alpha, Hunyuan-T1, LFM-40B, Mercury~Coder, and open-weight models (DeepSeek~R1, R1~distills at 14B/32B/70B, DeepSeek~V3, DeepSeek~Prover~V2, DeepCoder~14B, Qwen~3~235B/32B/14B/8B/4B/1.7B, Qwen~3~Coder~480B, Qwen~3~Next~80B, Qwen~2.5~72B/Coder~7B/14B/32B, QwQ-32B, Qwerky-72B, Llama~3.1~405B, 3.1~Tulu-3~405B, 3.2~90B~Vision, 3.3~70B, 4~Maverick, 4~Scout, Gemma~3, Phi-4, Mixtral~8$\times$7B, Mistral~Small~3.1~24B, Cogito~v2.1~671B, GLM-4.5~Air, GLM-4.7, MiniMax-01, Jamba~1.6~Large, OLMo-2, Prime~Intellect~3, GPT-OSS~20B/120B).

Models are accessed via their respective APIs. Each model is run once per graph. Where models support extended thinking or ``high'' reasoning effort, both settings are tested as separate entries.

\subsection{Execution}

Each run calls the model API with the prompt, extracts the last code block from the response, validates it, and records the result (pass/fail, duration, error details) to a SQLite database. Runs already recorded for a given (model, graph\_index) pair are skipped, enabling incremental evaluation.

\newpage
\section{Results}

\subsection{Leaderboard}

Table~\ref{tab:leaderboard} shows the top-performing models.

\begin{table}[h]
\centering
\caption{Top 21 models by score (of 199). Full results cover 91 models.}
\label{tab:leaderboard}
\small
\begin{tabular}{@{}rlr@{}}
\toprule
Score & Model & Time (s) \\
\midrule
159.5 & gpt-5.2-pro-medium & 80{,}028 \\
151.0 & gemini3-pro-preview & 8{,}956 \\
144.0 & opus-4-5-32k-stream & 29{,}292 \\
130.0 & gpt-5-codex & 106{,}930 \\
127.0 & gpt-5-codex-high & 98{,}787 \\
124.5 & gpt-5-2025-08-07 & 37{,}875 \\
124.0 & gpt-5-2025-08-07-high & 50{,}602 \\
112.5 & sonnet-4-5-32k-stream & 21{,}423 \\
111.0 & gpt-5-mini-2025-08-07-high & 50{,}982 \\
101.5 & gpt-5.2-low & 9{,}616 \\
96.0 & gemini-2.5-pro-preview-05-06 & 15{,}813 \\
95.5 & sonnet-3-7-0219-32k & 24{,}942 \\
93.5 & o4-mini & 11{,}622 \\
91.0 & o1 & 19{,}255 \\
85.0 & gemini-2.5-pro-preview-03-25 & 6{,}380 \\
84.0 & glm-4.7 & 224{,}486 \\
83.5 & haiku-4-5-32k-stream & 15{,}374 \\
82.0 & gpt-5-nano-2025-08-07-high & 28{,}716 \\
81.0 & o3 & 33{,}736 \\
80.0 & grok-4-1-fast & 34{,}905 \\
78.0 & k2-thinking & 45{,}220 \\
\bottomrule
\end{tabular}
\end{table}

\subsection{Key Findings}

\paragraph{Finding 1: Extended thinking and model scale both improve spatial reasoning.}
Sonnet~3.7 jumps from 39.5/199 without thinking to 95.5/199 with 32k thinking tokens---by far the largest gain from reasoning effort alone. Moonshot K2 shows a similar leap (25.0 $\to$ 78.0 with thinking), and GPT-5~Mini improves substantially (78.0 $\to$ 111.0 with ``high'' reasoning). Within a model family at fixed thinking budget, scale matters: Opus~4.5 (144.0) $>$ Sonnet~4.5 (112.5) $>$ Haiku~4.5 (83.5), all with 32k tokens. However, thinking does not help uniformly: GPT-5~``high'' (124.0) matches base GPT-5 (124.5), suggesting the benefit depends on model architecture.

\paragraph{Finding 2: Spatial reasoning does not track standard benchmark rankings.}
Claude Opus~4.5 (144.0/199) outperforms GPT-5 (124.5/199) on PlanarBench despite scoring lower on the Artificial Analysis Intelligence Index (43 vs.\ 45). The o1 model (91.0/199) underperforms Sonnet~3.7 with extended thinking (95.5/199). Standard leaderboard positions are poor predictors of spatial layout ability.

\paragraph{Finding 3: A capability cliff exists around 30B parameters.}
Models below 30B parameters almost universally score under 20/199: Phi-4 (13.5), Gemma~3 (12.0), Qwen~2.5~Coder~32B (17.0). The smallest models (7B range) score under 11. Spatial reasoning appears to require a minimum representational capacity that small models lack.

\paragraph{Finding 4: Edge count is the primary difficulty driver.}
The Pearson correlation between a task's edge count and its score (averaged across all models) is $r = -0.85$, 95\% CI $[-0.88, -0.80]$. By comparison, vertex count alone correlates at only $r = -0.47$ $[-0.57, -0.36]$---the confidence intervals do not overlap. After controlling for vertex count via partial correlation, edge count retains $r = -0.80$ $[-0.84, -0.74]$.

\begin{figure}[t]
\centering
\includegraphics[width=\linewidth]{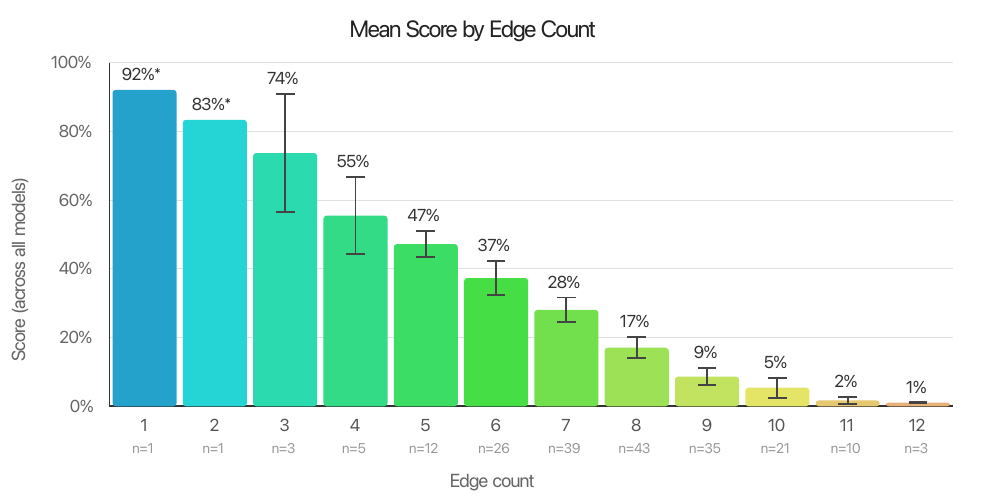}
\caption{Mean score (across all models) decreases monotonically with edge count. Error bars show 95\% CIs; asterisked bars ($n=1$) have no CI. Sample sizes shown below each bar.}
\label{fig:bar}
\end{figure}

\begin{figure}[t]
\centering
\includegraphics[width=\linewidth]{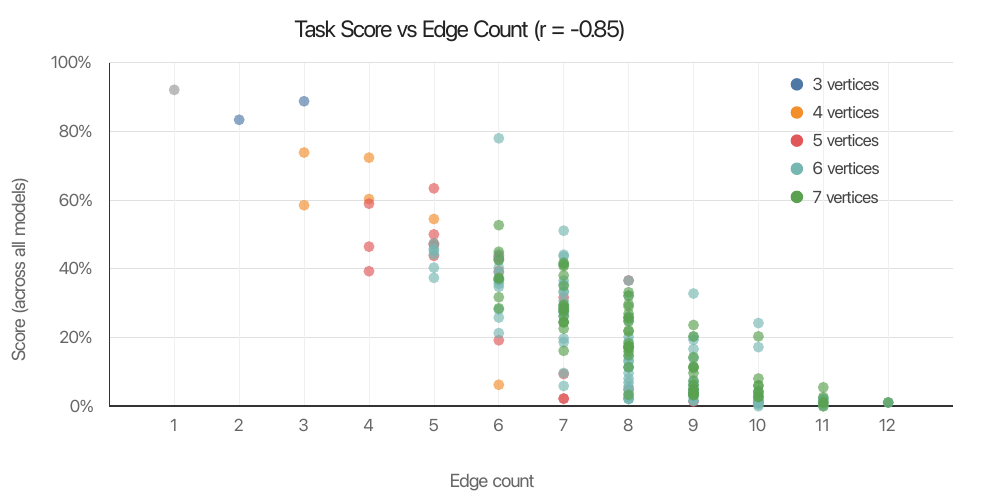}
\caption{Each point is one benchmark task, colored by vertex count. The strong negative correlation ($r = -0.85$) holds within every vertex group, confirming that edge density---not just graph size---drives difficulty.}
\label{fig:scatter}
\end{figure}

Table~\ref{tab:binned} shows the binned solve rates, and Figures~\ref{fig:bar} and~\ref{fig:scatter} visualize the relationship. Bins with $n = 1$ have no meaningful CI. The 3-edge bin ($n = 3$) has a wide interval; bins with $n \ge 5$ have CIs within $\pm$6 percentage points.

Even under relaxed scoring, the transition from ``usually solvable'' ($\le$5 edges) to ``rarely solved'' ($\ge$9 edges) spans a narrow range. Tasks with 11--12 edges see solve rates below 2\%, suggesting a near-hard threshold in current model capability.

\begin{table}[h]
\centering
\caption{Mean score by edge count (across all models). Bins with $n=1$ have no meaningful CI.}
\label{tab:binned}
\small
\begin{tabular}{@{}rrrr@{}}
\toprule
Edges & Tasks & Mean Score & 95\% CI \\
\midrule
1 & 1 & 92.0\% & --- \\
2 & 1 & 84.2\% & --- \\
3 & 3 & 73.9\% & [56.2\%, 91.6\%] \\
4 & 5 & 55.0\% & [43.6\%, 66.4\%] \\
5 & 12 & 46.4\% & [42.2\%, 50.7\%] \\
6 & 26 & 36.4\% & [31.5\%, 41.4\%] \\
7 & 39 & 27.1\% & [23.5\%, 30.7\%] \\
8 & 43 & 16.3\% & [13.3\%, 19.3\%] \\
9 & 35 & 8.3\% & [5.9\%, 10.7\%] \\
10 & 21 & 5.1\% & [2.3\%, 7.9\%] \\
11 & 10 & 1.6\% & [0.6\%, 2.5\%] \\
12 & 3 & 1.0\% & [1.0\%, 1.0\%] \\
\bottomrule
\end{tabular}
\end{table}

\paragraph{Finding 5: Task score is not associated with output size.}
One possible explanation for the edge-count association is that graphs with more
edges simply require larger or longer outputs. However, we found no meaningful association
between task score and either drawing area or total response length. The observed
edge-count relationship therefore does not appear to be reducible to the amount of
output models must generate.

\begin{figure}[t]
\centering
\includegraphics[width=\linewidth]{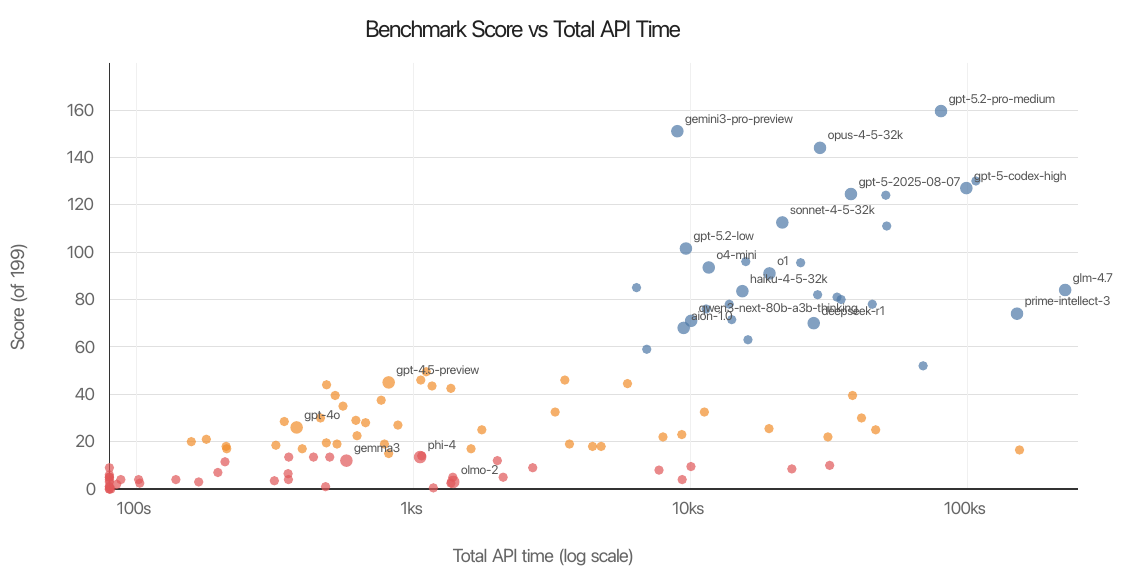}
\caption{Each point is one model. The $x$-axis (log scale) shows total API wall-clock time; the $y$-axis shows score. Colors indicate score tiers: blue (50+), orange (15--49), red ($<$15).}
\label{fig:time}
\end{figure}

\paragraph{Finding 6: Cost-efficiency varies enormously.}
Gemini~3~Pro~Preview achieves the second-highest score (151.0/199) in 8{,}956 seconds of total API time. GPT-5~Codex achieves a comparable score (130.0/199) but takes 106{,}930 seconds---over 11$\times$ slower. Prime Intellect~3 achieves 74.0/199 in 150{,}475 seconds. Speed and quality are not correlated (Figure~\ref{fig:time}).

\begin{figure}[t]
\centering
\includegraphics[width=\linewidth]{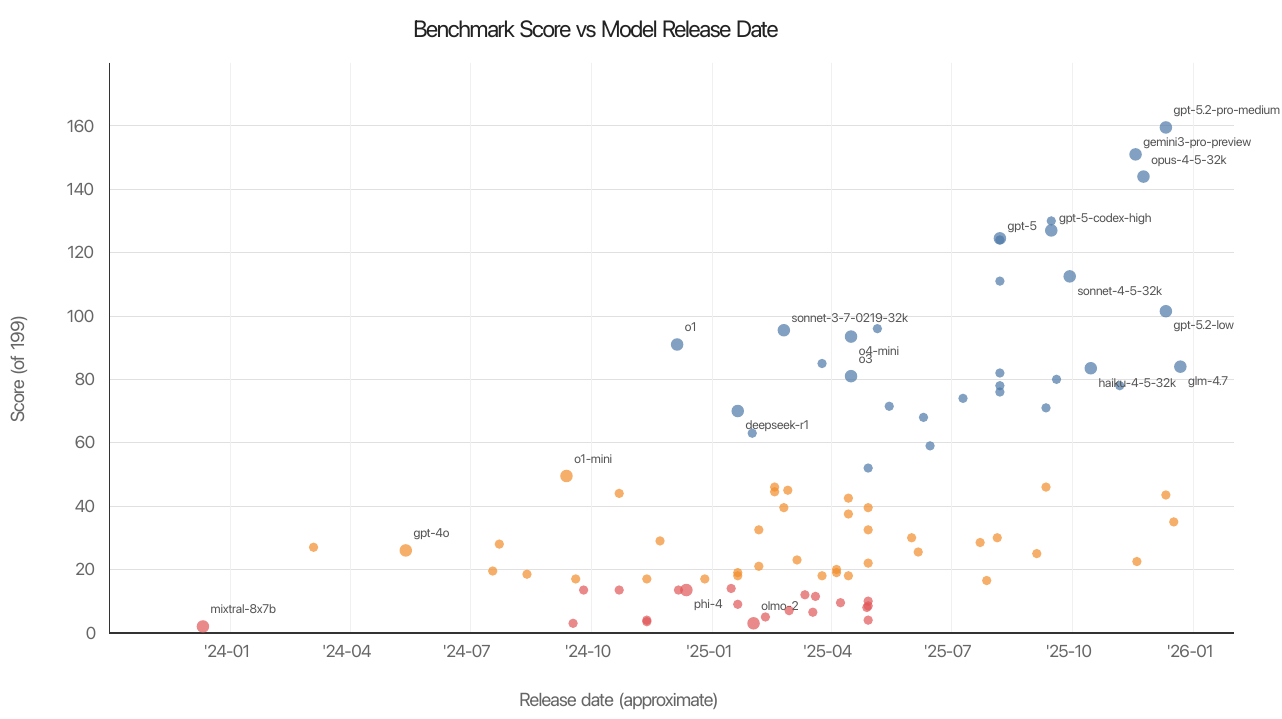}
\caption{Score vs model release date. Newer models score higher, and with the exception of o1 the frontier is almost linearly growing.}
\label{fig:release}
\end{figure}

\subsection{Error Analysis}

Common failure modes include:
\begin{itemize}[nosep]
\item \textbf{Missing edges:} The model places nodes correctly but fails to connect all required pairs---the most common failure, particularly for high-degree nodes.
\item \textbf{Extra edges:} The model draws connections not in the input graph, often because dash characters in ``decorative'' spacing are parsed as edges.
\item \textbf{Node placement failures:} Missing or duplicate node labels, particularly when models add annotations outside the intended drawing area.
\item \textbf{Broken edges:} Edge characters that don't form a contiguous path between their endpoints---a dash or pipe that is slightly misaligned.
\end{itemize}

\newpage
\section{Discussion}

\subsection{Why ASCII Art?}

We chose ASCII art over coordinate-pair output for several reasons. First, ASCII art is a practical output modality---it is widely used for diagrams and documentation, and LLMs encounter it in training data. Second, it tightly couples node placement with edge routing: the model must simultaneously reason about where nodes go and whether edges can be drawn between them without crossing.

The downside is that ASCII art constrains the space of valid drawings. Diagonal edges must use \texttt{/} and \texttt{\textbackslash} characters, which limits angular resolution. Some planar embeddings (e.g.\ the complete $K_4$ graph) that would be easy to draw on a continuous 2D canvas are hard to express in ASCII. To account for this, our scoring awards partial credit for drawings with correct node placement or correct edge connectivity via non-straight paths (see Section~3.3). This arguably makes the benchmark \emph{harder} than coordinate-based alternatives, which we view as acceptable---a benchmark where the best model scores 159.5 of 199 is more informative than one they solve nearly all of. The open-ended design of the benchmark was inspired by AidanBench~\citep{mclaughlin2025aidanbench}, and the idea originally arose in the context of LLM-assisted requirements engineering, where we observed a correlation between a model's ability to render dependency diagrams correctly and its ability to spot contradictions between task statements.

\subsection{Edge Density as a Difficulty Axis}

Vertex count alone is a weak predictor of difficulty (Pearson $r = -0.47$), while edge count dominates ($r = -0.85$). The partial correlation of edge count with score, controlling for vertices, remains strong at $r = -0.80$ $[-0.84, -0.74]$.

Each additional edge constrains the layout more tightly, reducing the space of valid planar embeddings. A 7-vertex tree (6~edges) is far easier to draw than a 6-vertex maximal planar graph (12~edges), despite having more nodes. The combinatorial bottleneck is not ``where to put the nodes'' but ``how to route all edges without crossings given the node positions.''

Even with relaxed scoring that recognizes non-straight edge connectivity and edges skipped due to acute angles, tasks with 11--12 edges see scores below 2\%, suggesting that current models hit a near-hard limit in their spatial reasoning capacity at this edge density.

\subsection{What Does the Edge-Count Effect Measure?}

Vertex count approximates the number of entities that must be represented, whereas edge count approximates the number of relations that must be realized jointly. The substantially stronger association with edge count is therefore consistent with a relational-load account of task difficulty: performance deteriorates not primarily because more objects must be placed, but because more pairwise requirements must be satisfied within one globally consistent construction. The absence of meaningful associations with drawing area and response length makes a simple output-volume explanation less plausible.

This entity--relation distinction is conceptually related to relational-complexity
accounts of human reasoning, which associate processing demand with the relations
that must be represented jointly \citep{halford1998}. The analogy is necessarily
limited, however: classical relational complexity is defined primarily by relational
arity, rather than by the raw number of pairwise relations.

This interpretation also highlights a distinction between query answering and constructive evaluation. Even when answering a graph-theoretic query requires global computation, the model usually returns a short answer, path, or scalar value. A planar drawing instead constitutes a complete witness: every required adjacency must be realized, no prohibited adjacency may be introduced, and all placements and routes must remain mutually compatible. Because generation is sequential, early placement decisions may constrain later edge routes and create a need for revision or backtracking. PlanarBench may therefore expose limitations in constructing globally consistent outputs that remain hidden when only the final answer to a graph query is evaluated.

However, edge count also increases the number of opportunities for local failure. If each required edge were rendered correctly with independent probability $p$, the probability of producing an entirely correct $m$-edge drawing would decrease approximately as $p^m$, even if the edges imposed no additional cognitive interaction. The present results therefore establish edge count as a strong empirical difficulty axis, but do not yet determine how much of the effect reflects relational integration, accumulated local errors, or graph properties correlated with edge count. The output-size analysis argues against a crude verbosity account, but does not eliminate this conjunctive-error explanation.

Distinguishing these mechanisms will require finer-grained scoring in future studies. .

\subsection{Limitations}

\begin{itemize}[nosep]
\item \textbf{Single-attempt evaluation.} Each model gets one attempt per graph. Some models might succeed on retry with different sampling. Broader multi-attempt evaluation was constrained by the API costs of frontier reasoning models. Future work could evaluate pass@$k$.
\item \textbf{ASCII-specific failures.} Some failures reflect difficulty with the ASCII medium rather than spatial reasoning per se. The relaxed scoring (coordinate and BFS checks) partially addresses this by awarding credit when topology is correct but edge paths are non-straight.
\item \textbf{Graph size.} The current benchmark maxes out at 7 vertices. However, the benchmark is open-ended by design: McKay's corpus contains 71{,}885 planar graphs at 9 vertices and the counts grow super-exponentially, so the task set can be extended indefinitely as models improve. Scaling beyond ASCII would require switching to a coordinate-output format.
\item \textbf{Prompt sensitivity.} We use a single, simple prompt. Prompt engineering (e.g., chain-of-thought, few-shot examples) might improve results but was not explored.
\end{itemize}

\section{Conclusion}

Planar graph drawing is a discriminating test of LLM capabilities: simple to state, easy to verify, difficult to solve via memorization, and it exposes a wide performance spread across models. Edge density appears to be an important difficulty driver ($r = -0.85$), with scores below 2\% for tasks with 11+ edges even under relaxed scoring. Prior major LLM graph benchmarks (GraphArena, NLGraph, GraphOmni) use node count as the main difficulty axis. However, our results show that edge count is a stronger predictor and remains dominant after controlling for vertex count. Extended thinking and model scale are the dominant predictors of success, and the ranking on this spatial task diverges meaningfully from standard LLM leaderboards.

PlanarBench is open-ended: since the number of planar graphs grows super-exponentially with vertex count, the benchmark can be extended indefinitely to challenge more powerful future models without redesigning the task or evaluation. We release the graph corpus and evaluation harness to support further research on spatial reasoning in language models: \url{https://github.com/wizzard0/planar-bench-as1073}

For future work our priority is testing repeated runs for each model to estimate instance-level success probabilities and pass@$k$, distinguishing persistent capability limits from stochastic decoding failures. Open-weight models could also be compared as a function of both total and active parameter count, particularly for mixture-of-experts architectures, to systematically evaluate how decoding and sampling strategies affect performance and reliability. Another important direction for future validation is permutation robustness. 

An important theoretical direction is also to separate the mechanisms underlying the edge-count association. A balanced design sampling multiple graphs within fixed vertex- and edge-count strata would permit comparisons of cycle rank, degree structure, connectivity, embedding flexibility, and other candidate difficulty measures.

\newpage
\bibliographystyle{plainnat}

\begin{thebibliography}{20}

\bibitem[Babai(2016)]{babai2016}
L.~Babai.
\newblock Graph isomorphism in quasipolynomial time.
\newblock In \emph{Proceedings of the 48th Annual ACM Symposium on Theory of Computing (STOC)}, pp.\ 684--697, 2016.

\bibitem[Brinkmann \& McKay(2007)]{brinkmann2007}
G.~Brinkmann and B.~D. McKay.
\newblock Fast generation of planar graphs.
\newblock \emph{MATCH Communications in Mathematical and in Computer Chemistry}, 58(2):323--357, 2007.

\bibitem[de~Fraysseix et~al.(1990)]{defraysseix1990}
H.~de~Fraysseix, J.~Pach, and R.~Pollack.
\newblock How to draw a planar graph on a grid.
\newblock \emph{Combinatorica}, 10(1):41--51, 1990.

\bibitem[Di~Bartolomeo et~al.(2023)]{dibartolomeo2023}
S.~Di~Bartolomeo, G.~Severi, V.~Schetinger, and C.~Dunne.
\newblock Ask and you shall receive (a graph drawing): Testing {ChatGPT}'s potential to apply graph layout algorithms.
\newblock In \emph{EuroVis 2023 Short Papers}, 2023.

\bibitem[Fan et~al.(2025)]{fan2025}
Y.~Fan, X.~Lyu, L.~Wang, Y.~Zhao, F.~Zhou, and Y.~Wang.
\newblock How well will {LLM}s perform for graph layout tasks?
\newblock \emph{Visual Informatics}, 100285, 2025.

\bibitem[F\'ary(1948)]{fary1948}
I.~F\'ary.
\newblock On straight-line representation of planar graphs.
\newblock \emph{Acta Scientiarum Mathematicarum (Szeged)}, 11(4):229--233, 1948.

\bibitem[Hopcroft \& Wong(1974)]{hopcroft1974}
J.~E. Hopcroft and J.~K. Wong.
\newblock Linear time algorithm for isomorphism of planar graphs (preliminary report).
\newblock In \emph{Proceedings of the 6th Annual ACM Symposium on Theory of Computing (STOC)}, pp.\ 172--184, 1974.

\bibitem[McKay(2024)]{mckay2024}
B.~D. McKay.
\newblock Combinatorial data: Planar graphs.
\newblock Australian National University, 2024.
\newblock \url{https://users.cecs.anu.edu.au/~bdm/data/graphs.html}.

\bibitem[McKay \& Piperno(2014)]{mckay2014}
B.~D. McKay and A.~Piperno.
\newblock Practical graph isomorphism, {II}.
\newblock \emph{Journal of Symbolic Computation}, 60:94--112, 2014.

\bibitem[McLaughlin et~al.(2024)]{mclaughlin2025aidanbench}
A.~McLaughlin, J.~Campbell, A.~Uppuluri, and Y.~Yang.
\newblock {AidanBench}: Stress-testing language model creativity on open-ended questions.
\newblock In \emph{NeurIPS 2024 Workshop on Language Gamification}, 2024.
\newblock \url{https://github.com/aidanmclaughlin/AidanBench}.

\bibitem[Tang et~al.(2025)]{tang2025}
J.~Tang, Q.~Zhang, Y.~Li, N.~Chen, and J.~Li.
\newblock {GraphArena}: Evaluating and exploring large language models on graph computation.
\newblock In \emph{International Conference on Learning Representations (ICLR)}, 2025.

\bibitem[Tutte(1963)]{tutte1963}
W.~T. Tutte.
\newblock How to draw a graph.
\newblock \emph{Proceedings of the London Mathematical Society}, s3-13(1):743--767, 1963.

\bibitem[Wang et~al.(2023)]{wang2023}
H.~Wang, S.~Feng, T.~He, Z.~Tan, X.~Han, and Y.~Tsvetkov.
\newblock Can language models solve graph problems in natural language?
\newblock In \emph{Advances in Neural Information Processing Systems (NeurIPS)}, 2023.

\bibitem[Xu et~al.(2026)]{xu2026}
H.~Xu, X.~Jian, X.~Zhao, W.~Pang, C.~Zhang, S.~Wang, Q.~Zhang, Z.~Dong, J.~Monteiro, B.~Liu, Q.~Sun, and T.~Yu.
\newblock {GraphOmni}: A comprehensive and extensible benchmark framework for large language models on graph-theoretic tasks.
\newblock In \emph{International Conference on Learning Representations (ICLR)}, 2026.

\bibitem[Li et~al.(2024)]{li2024stepgame}
F.~Li, D.~C. Hogg, and A.~G. Cohn.
\newblock Advancing spatial reasoning in large language models: An in-depth evaluation and enhancement using the {StepGame} benchmark.
\newblock In \emph{Proceedings of the AAAI Conference on Artificial Intelligence (AAAI)}, 38(17):18500--18507, 2024.

\bibitem[Wu et~al.(2024)]{wu2024mindseye}
W.~Wu, S.~Mao, Y.~Zhang, Y.~Xia, L.~Dong, L.~Cui, and F.~Wei.
\newblock Mind's eye of {LLM}s: Visualization-of-thought elicits spatial reasoning in large language models.
\newblock In \emph{Advances in Neural Information Processing Systems (NeurIPS)}, 2024.

\bibitem[Premsri \& Kordjamshidi(2025)]{premsri2025forest}
T.~Premsri and P.~Kordjamshidi.
\newblock {FoREST}: Frame of reference evaluation in spatial reasoning tasks.
\newblock In \emph{Proceedings of the 2025 Conference on Empirical Methods in Natural Language Processing (EMNLP)}, pp.\ 34977--35003, 2025.

\bibitem[Ge et~al.(2024)]{ge2024}
Y.~Ge, S.~Liu, W.~Feng, L.~Mei, L.~Chen, and X.~Cheng.
\newblock Graph descriptive order improves reasoning with large language model.
\newblock \emph{arXiv preprint arXiv:2402.07140}, 2024.

\bibitem[Heyman \& Zylberberg(2025)]{heyman2025}
A.~Heyman and J.~Zylberberg.
\newblock Evaluating the systematic reasoning abilities of large language models through graph coloring.
\newblock \emph{arXiv preprint arXiv:2502.07087}, 2025.

\bibitem[Halford et~al.(1998)]{halford1998}
G.~S. Halford, W.~H. Wilson, and S.~Phillips.
\newblock Processing capacity defined by relational complexity:
Implications for comparative, developmental, and cognitive psychology.
\newblock \emph{Behavioral and Brain Sciences}, 21(6):803--831, 1998.

\end{thebibliography}

\newpage
\appendix
\section{Relaxed Validation Examples}
\label{app:examples}

The following examples illustrate drawings that fail strict validation but pass one or both relaxed checks. Each category is named by its (strict, coordinate, BFS) pass/fail pattern.

\subsection*{Category 0-0-1: BFS only}

These drawings fail both strict and coordinate checks but pass BFS validation, typically because an edge is routed around an intervening node rather than drawn as a straight line.

\paragraph{Task 17 (5 nodes).} Coordinate check rejects edge D--E because the straight line passes through B; BFS recovers the outside route.

\begin{verbatim}
           D---------+
          /|\        |
         / | \       |
        A  B  C      |
         \ | /       |
          \|/        |
           E---------+
\end{verbatim}

\paragraph{Task 126 (6 nodes).} Coordinate check rejects A--B by straight-line intersection; BFS accepts the outer routed path.

\begin{verbatim}
                 A-----------+
                / \          |
               /   \         |
              /     \        |
             C--------E      |
             |      / |      |
             |     /  |      |
             |    /   |      |
             |   /    |      |
             |  /     |      |
             | /      |      |
             |/       |      |
             F--------D      |
              \      /       |
               \    /        |
                \  /         |
                 B-----------+
\end{verbatim}

\paragraph{Task 159 (7 nodes).} Coordinate check rejects E--F because the straight segment passes through C; BFS accepts the routed bottom edge.

\begin{verbatim}
    A       B
     \     /
      G   F
     /|\ /|
    D | C |
     \|   |
      E---'
\end{verbatim}
\newpage
\subsection*{Category 0-1-0: Coordinate only}

These drawings have valid node placements (all desired edges can be drawn as non-intersecting straight lines), but BFS cannot find a drawn path for every edge.

\paragraph{Task 22 (5 nodes).} Coordinates are acceptable, but BFS does not find D--E as a drawn edge.

\begin{verbatim}
       A
      / \
     C - E
    /     \
   D-------B
\end{verbatim}

\paragraph{Task 62 (6 nodes).} Coordinates are acceptable, but BFS misses C--F.

\begin{verbatim}
        A
        |
        F
       /|\ \
      / | \ \
     /  |  \ \
    /   |   \ \
   B----D----E----C
\end{verbatim}

\paragraph{Task 187 (7 nodes).} Coordinates are acceptable, but BFS misses F--G and C--E.

\begin{verbatim}
          A
          |
          |
          |
          |
    C-----G-----B
    |    /|\
    |   / | \
     \ /  |  \
      E   |   D
      |    \ /
      +-----F
\end{verbatim}
\newpage
\subsection*{Category 0-1-1: Coordinate and BFS}

These drawings pass both relaxed checks but fail strict validation, typically because edges use L-shaped or routed paths rather than straight-line ASCII characters.

\paragraph{Task 3 (3 nodes).} BFS recovers the routed B--C edge via the corner.

\begin{verbatim}
A---B
|   |
C---+
\end{verbatim}

\paragraph{Task 5 (4 nodes).} Can be drawn with straight lines, but needs BFS to recognize paths for A--D and C--D.

\begin{verbatim}
A---+
    |
    D---B
    |   |
    |   |
    +---C
\end{verbatim}

\paragraph{Task 166 (7 nodes).} Can be drawn with straight lines, but needs BFS to recognize the angled D--E and E-G edges.

\begin{verbatim}
      A
      |
D-----F-----G
|     |    /|\
|     |   / | \
+-----E--/  C  B
\end{verbatim}

\end{document}